\documentclass[conference]{IEEEtran}
\IEEEoverridecommandlockouts
\usepackage{cite}
\usepackage{amsmath,amssymb,amsfonts}
\usepackage{algorithmic}
\usepackage{graphicx}
\usepackage{textcomp}
\usepackage{hyperref}
\usepackage{tikz,pgfplots}
\pgfplotsset{width=10cm,compat=1.9}
\usetikzlibrary{calc}

\usepackage{pgfplotstable}
\usetikzlibrary{positioning}
\usetikzlibrary{pgfplots.groupplots}
\usetikzlibrary{arrows}

\usetikzlibrary{arrows}

\usepackage{xcolor,mathcomSTEv4}
\def\BibTeX{{\rm B\kern-.05em{\sc i\kern-.025em b}\kern-.08em
    T\kern-.1667em\lower.7ex\hbox{E}\kern-.125emX}}

\newcommand{\topK}{ {\rm top} K}

\newcommand{\comp}{ {\rm comp}}

\begin{document}

\title{
Lossy Gradient Compression: \\
How Much Accuracy Can One Bit Buy?
}

\author{\IEEEauthorblockN{Sadaf Salehkalaibar}
\IEEEauthorblockA{\textit{Electrical and Computer Engineering Department} \\
\textit{McMaster University}\\
salehkas@mcmaster.ca}
\and
\IEEEauthorblockN{Stefano Rini}
\IEEEauthorblockA{\textit{Electrical and Computer Engineering Department} \\
\textit{National Yang-Ming Chiao-Tung University (NYCU)}\\
stefano.rini@nycu.edu.tw}
%
}

\maketitle

\begin{abstract}
In federated learning (FL), a global model is trained at a Parameter Server (PS) by aggregating model updates obtained from multiple remote learners.
%
Generally, the communication between the remote users and the PS is rate-limited,
while the transmission from the PS to the remote users are unconstrained. 
The FL setting gives rise to the distributed learning scenario in which the updates from the remote learners have to be compressed so as to meet communication rate constraints in the uplink transmission toward the PS. 
For this problem, one wishes  to compress the model updates so as to minimize the loss in accuracy resulting from the compression error.
In this paper, we take a rate-distortion approach to address the compressor design problem for the distributed training of deep neural networks (DNNs).
%
%
In particular, we define a measure of the compression performance under communication-rate constraints-- the \emph{per-bit accuracy}-- which addresses the ultimate improvement of  accuracy that a bit of communication brings to the centralized model.
In order to maximize the per-bit accuracy, we consider  modeling the 
DNN gradient updates at remote learners
%
as  a generalized normal distribution. 
Under this assumption on the DNN gradient distribution, we propose a class of distortion measures to aid  the design of quantizers for the compression of the model updates.
We argue that this family of distortion measures, which we refer to as ``$M$-magnitude weighted $L_2$'' norm, captures the practitioner's intuition in the choice of gradient compressor. 
%
%
Numerical simulations are provided to validate the proposed 
approach for the CIFAR-10 dataset.
%
\end{abstract}


\section{Introduction}
Federated learning (FL) holds the promise of enabling the distributed training of large models over massive datasets while preserving data locality, guarantying scalability, and also preserving data privacy.
Despite the great advantages promised by the FL, the communication overhead of distributed training poses a challenge to present-day network.
As the size of the trained models  and the number of devices participating to the training is ever increasing, the transmission from remote users to the parameter server (PS) orchestrating the training process becomes the critical performance bottleneck  \cite{konevcny2016federated,li2019federated}.
In order to address this issue, the design of an effective gradient compressor is of fundamental importance. 
In this paper, we propose the \emph{per-bit accuracy} as a rate-distortion inspired approach to the  design of an efficient gradient compressor for FL. 
We also 
introduce a class of distortion measures for the design of gradient compression for the  distributed DNN training. 

\smallskip

\noindent
 {\bf Literature Review:}
FL consists of a central model which is trained   locally at the remote clients by applying Stochastic Gradient Descent (SGD) over a local dataset.
The local gradients are then communicated to the central PS for aggregation into a global model.
Since this aggregation model does not require data centralization, it provides substantial advantages in terms of scalability, robustness, and security. 
For these reasons, in the recent years, there has been a significant interest in developing effective and efficient FL algorithms  \cite{Shalev-Shwartz2010FL_CE,Wang2018Spars_FL,Alistarh2018Spars_FL}.
%

 A natural constraint in distributed and decentralized optimization is with respect to transmission rates between nodes and its relationship to the overall accuracy \cite{shlezinger2020communication}.
%
Common in these approaches is the effort put forth in reducing the dimensionality of the gradients.
Following \cite{SGD_sparse2}, we shall refer to these dimensionality reduction schemes as \emph{gradient compressors}.
Gradient compressors can be divided in two classes:  (i) gradient sparsification, \cite{Shalev-Shwartz2010FL_CE,Wang2018Spars_FL,Alistarh2018Spars_FL}, 
and (ii) gradient quantization  \cite{seide2014onebitSGD,Konecny2016Fl_CE, gandikota2019vqsgd,Bernstein2018signSGC,FL_DSGD_binomial,Li2019DP_CEFL}. 
These compression schemes improve the communication efficiency by reducing the dimensionality of the message being transmitted from the users to the PS through (i) sparsifying the transmitted model update
and (ii) stochastic scalar quantization, i.e.  non-subtractive dither quantization. 
In some literature, the dimensionality-reduction is performed on the whole gradient vector: for instance  the authors of \cite{gandikota2019vqsgd} introduce vector  quantization for SGD.

\smallskip

\noindent
 {\bf Contribution:}
In the following, we shall consider the problem of rate-efficient gradient compression for distributed training.
We consider, in particular,  the FL training of DNNs and present relevant numerical simulations for this case.
We introduce the \emph{per-bit accuracy} as the relevant performance measure for distributed training under communication constraints.
The per-bit accuracy corresponds to the  improvement in accuracy that a gradient compressed within $\Rsf$ bits can provide to a given model.
Next, we consider a we propose a rate-distortion  approach  to maximize the per-bit accuracy.
In this approach, the quantizer design results from two components (i) the gradient distribution, and (ii) the choice of distortion measure between the original and the compressed gradient. 
For (i), we follow the modeling of the DNN gradients as in \cite{chen2021dnn}  in which it is argued that the DNN gradients can be effectively modelled as a generalized normal distribution (GenNorm). We argue tat 
For (ii), we propose a set of distortion measures between the original and the compressed gradient in which the $L_2$ loss is further weighted by the magnitude of the gradient to the power of $M$. 
The paper concludes with numerical simulation which show that a  compressor designed through this approach outperforms other compressors considered in the literature, such as the gradient quantization and sparsification. 
%

\smallskip
\noindent
{\bf Notation:} In the following, lower case boldface letters (eg. $\zv$) are used for column vectors and uppercase boldface letters (eg. $\Mv$) designate matrices. 
%
%
The all-zero vector of dimension $d$ is indicated as $\zerov_d$.
We also adopt the shorthands $[m:n] \triangleq \{m, \ldots, n\}$
and  $[n] \triangleq \{1, \ldots, n\}$. 
The $p$-norm of the vector $\xv$ is indicated as $\| \xv\|_p$.
%
%
%
Calligraphic scripts are used to denote sets (e.g. $\Acal$) and $|\Acal|$ is used to denote its cardinality.

\section{System Model}
\label{sec:System Model}
In the following, we consider the distributed training of a machine learning (ML) model across $N$ devices where the communication between the remote device and the PS is limited to $R$ bits per learner. 
Let us begin by introducing the optimization and the communication setting, followed by the formal statement of our performance optimization problem.

\subsection{Distributed Optimization Setting}
\label{sec:Optimization Setting}

Consider the scenario with $N$ clients each posses a local dataset $\Dcal_n=\{ \dv_{nk} \}_{k \in [|\Dcal_n|] }$ for $n \in [N]$ and wish to minimize the \emph{loss function} $\Lcal$ as evaluated across all the local datasets and over the choice of model $\wv \in \Rbb^d$, that is 
\ea{
\Lcal(\wv) =  \f 1 {\sum_{n \in [N]} |\Dcal_n|} \sum_{n \in [N]} \sum_{\dv_{nk} \in [\Dcal_n]} \Lcal(\dv_{nk}, \wv).
\label{eq:loss}
}
%
For the  {loss function} in \eqref{eq:loss}, we assume that there exists a unique minimizer $\wv^*$ of \eqref{eq:loss}, that is, 
\ea{
\wv^*= \argmin_{\wv \in \Rbb^d} \ \ \Lcal(\wv).
\label{eq:minimum w}
}
A common approach for numerically determining the optimal value in \eqref{eq:minimum w} in the centralized scenario is through the iterative application of (synchronous) stochastic gradient descent (SGD).
In the (centralized) SGD algorithm, the learner maintains a  estimate of the minimizer in \eqref{eq:minimum w}, $\wv_t$, for each time $t\in[T]$. The final estimate of \eqref{eq:minimum w} is $\wv_T$. 
At each time $t\in[T]$, the estimate $\wv_t$ is updated as
%
%
\ea{
\wv_{t+1}=\wv_{t}-\eta_t  \gv_t,
\label{eq:SGD}
}
for $\wv_{0}=\zerov_d$, where $\eta_t$ is  an iteration-dependent step size $\eta_t$ called \emph{learning rate}, and where  $\gv_t$ is the stochastic gradient of $\Lcal$ evaluated in $\wv_{t}$, that is 
\ea{
\Ebb\lsb \gv_t\rsb= \sum_{ \dv_k \in \Dcal } \nabla\Lcal\lb \dv_k, \wv_t\rb.
\label{eq:mean gradient}
}
In \eqref{eq:mean gradient}, $\nabla\Lcal\lb\wv_n\rb$ denotes the gradient of $\Lcal\lb\\dv_k,wv_t\rb$ in $\wv_t$ as evaluated over the dataset  $\Dcal = \bigcup_{n \in [N]} \Dcal_n$.
%
%
%
%
%

In the FL setting, given that the datasets $\Dcal_{n}$ are distributed at multiple remote learners, the SGD algorthim as in \eqref{eq:SGD}  has to be adapted as follows. 
%
%
%
%
First (i) the PS  transmits the current model estimate, $\wv_t$,
to each client $n \in[N]$,
%
%
then (ii) each  client $n \in [N]$ accesses its  local dataset $\Dcal_{n}=\left\{\lb \dv_{n}(k),v_{n}(k) \rb\right\}_{k\in\left[\left|\Dc_{n}\right|\right]}$ and computes the stochastic gradient, $\gv_{nt}$, as in \eqref{eq:mean gradient} and communicates it to the PS.
Finally (iii) the PS updates the model estimate as  in \eqref{eq:SGD} but where $\gv_t$ is obtained as
\ea{
\gv_t=\f{1}{N}\sum_{n \in[N]}\gv_{nt},
\label{eq:aggregate}
}
and uses $\gv_t$ to update the model estimate as in \eqref{eq:SGD}.
We refer to the distributed version of SGD for the FL setting as \emph{federated averaging}.

\subsection{Federated Learning with Communication Constraints}
%
Customarily, in the FL setting, the communication is assumed to take place over some noiseless, infinity capacity link connecting the PS and the remote users and vice-versa. 
In a practical scenario, the users model  wireless mobiles, IoT devices, or sensors which have significant limitations in the available power and computational capabilities.
%
In these scenarios, we can still assume that  users rely on some physical and MAC layers' protocols that are capable of reliably delivering a certain payload from the users to the PS. 

For this reason, in the following, we assume that the communication between each of the users and the PS takes place over a rate-limited channel of capacity $d \Rsf$, where $d$ is the dimension of the model in Sec. \ref{sec:Optimization Setting}.
In other words, each client can communication up to $d \Rsf$ bits for each iteration $t \in [T]$.
In the following, we refer to the operation of converting the $d$-dimensional gradient vector $\gv_{nt}$ to a $d \Rsf$ binary vector as \emph{compression} and 
%
is indicated though the operator
\ea{
\comp_{\Rsf}: \ \   \Rbb^d \goes  [2^{d\Rsf}].
\label{eq:comp}
}
The reconstruction of the gradient is denoted by $\comp_{\Rsf}^{-1}$.
Note that in \eqref{eq:comp}, $\Rsf$ indicates the number of bits per model dimension. 
For simplicity, we assume in the following that all users are subject to the same communication constraint and all employ the same set of compressors. 

When gradients are compressed before transmissions, the evolution in \eqref{eq:SGD} is rewritten as
\ea{
\whv_{t+1}& =\whv_{t}-\eta_t  \ghv_t \nonumber \\
\ghv_t & = \f 1 n \sum_{n \in [N]} \comp_\Rsf^{-1} \lb\comp_\Rsf (\gv_{tn}) \rb.
\label{eq:SGD comp}
}


%

\subsection{Compression Performance Evaluation}
\label{sec:Compression Performance Evaluation}
In the following, we are interested in characterizing the compression performance in the terms of the loss of accuracy as a function of the communication rate.
More formally, give the model estimate $\whv_t$  and the gradient estimate $\ghv_t$, we wish determine 
\ea{
\Gsf_\Rsf(\whv_{t+1}) = \min_{\comp_{\Rsf},\comp_{\Rsf}^{-1}} \Lcal(\whv_{t+1}),
\label{eq:G}
}
where $\whv_{t+1}$ is obtained as in \eqref{eq:SGD comp}.
%
In general, we are interested in determining the effect of compression through the SGD iterations, so that 
$\Lcal(\wv_T) - \Gsf_{\Rsf}(\whv_{T})$ corresponds to the overall loss of accuracy due to compression at the training horizon, $T$.
%

The minimization in \eqref{eq:G} is generally too complex, as the loss function $\Lcal$ is generally non-convex in the model $\wv$. 
Additionally, lacking  a statistical description of the SGD  process is available, one cannot properly resort to classical compression techniques from information theory. 
To address these difficulties, later in Sec.  \ref{sec:A Rate-distortion Approach}, we consider a rate-distortion approach in which we simplify  the minimization in \eqref{eq:G} for the case of DNN training by considering (i) a family of  distortion measures which captures the loss of accuracy as a function of the gradient magnitude, and (ii) assume that the DNN gradients are iid  draws from the GenNorm distribution. 
These two simplifications yield a compressor design which shows improved performance over other compressors considered in the literature.



\subsection{DNN training}
\label{sec:DNN training}
While the problem formulation in Sec. \ref{sec:Compression Performance Evaluation} is rather general, in the remainder of the paper, we shall only consider the scenario of DNN training. 
%
%
More specifically, we  consider a DNN for classification of the CIFAR-10 dataset. 
The DNN model we consider is specified  in Table \ref{tab:model_layers}: the model is trained using SGD with a learning rate  $0.0001$, and cross-entropy as loss. 
The images are passed through the model in mini batches of $32$ at three epochs for every communication round. 
\begin{table}[t]
\vspace{0.5cm}
\footnotesize
    \centering
        \caption{A summary of the model layers of the DNN in Sec. \ref{sec:DNN training}. }
    \label{tab:model_layers}
\begin{tabular}{|c|c|c|}
\hline
Layer (Type)      & Output Shape    & Param No.            \\ \hline
conv2d (Conv2D)     & (None, 32, 32, 32)   & 896      \\ (\emph{top layer, \#6}) & & \\ \hline
batch\_normalization        &     (None, 32, 32, 32)         & 128                                      \\ \hline
conv2d\_1 (Conv2D)     & (None, 32, 32, 32)   & 9248        \\ \hline
batch\_normalization\_1       &   (None, 32, 32, 32)           & 128                                           \\ \hline
max\_pooling2d (MaxPooling2D)     & (None, 16, 16, 32)   & 0        \\ \hline
dropout (Dropout)     &   (None, 16, 16, 32)          & 0                                          \\ \hline
conv2d\_2 (Conv2D) & (None, 16, 16, 64)          & 18496 \\\hline
batch\_normalization\_2 & (None, 16, 16, 64) & 256 \\\hline
conv2d\_3 (Conv2D) & (None, 16, 16, 64) & 36928\\\hline
batch\_normalization\_3 & (None, 16, 16, 64) & 256\\\hline
max\_pooling2d\_1 & (None, 8, 8, 64) & 0 \\\hline
dropout\_1 (Dropout)  & (None, 8, 8, 64) & 0 \\\hline
conv2d\_4 (Conv2D)   & (None, 8, 8, 128) & 73856\\ (\emph{middle layer, \#24}) & & \\\hline
batch\_normalization\_4 & (None, 8, 8, 128) & 512\\\hline
conv2d\_5 (Conv2D) &  (None, 8, 8, 128) & 147584\\\hline
batch\_normalization\_5 & (None, 8, 8, 128) & 512\\\hline
max\_pooling2d\_2 & (None, 4, 4, 128) & 0\\\hline
dropout\_2 (Dropout) & (None, 4, 4, 128) & 0\\\hline
flatten (Flatten) & (None, 2048) & 0\\\hline
dense (Dense) & (None,128) & 262272\\\hline
batch\_normalization\_6 & (None, 128) & 512 \\\hline
dropout\_3 (Dropout) & (None,128) & 0 \\\hline
dense\_1 (Dense)  & (None,10) & 1290\\ (\emph{bottom layer, \#42}) & & \\\hline
\end{tabular}
\end{table}

\section{A Rate-distortion Approach to DNN Gradient Compression }
\label{sec:A Rate-distortion Approach}
%
In this section we introduce the  a rate-distortion approach for the design of the optimal compressor for DNN gradients. 
%
%
%
Since \eqref{eq:G} is generally intractable, we instead simplify the problem as follows. 
We choose (i) a certain distribution to approximate the gradient entries, and (ii) choose a distortion  that correlates the loss in accuracy in \eqref{eq:G}  when we compress the  original weights.
Once these two elements have been selected -- that a gradient distribution and a gradient distortion measure-- 
 the compressor in \eqref{eq:comp}  is chosen as the  the quantizer which minimize the chosen distortion for the given gradient distribution, as in the classic \cite{linde1980algorithm}.

\subsection{DNN Gradient Distribution}
\label{sec:DNN Gradient Distribution}

As argued in \cite{chen2021dnn}, the gradient entries observed performing the DNN training are well approximated  by the generalized normal distribution \cite{varanasi1989parametric}.
Note that the generalized normal distribution encompasses the Laplace and normal distribution as a special case: DNN gradients have been assumed to follow a Laplace distribution in \cite{isik2021successive} while they have been assumed to have a normal distribution in \cite{lee2017deep,matthews2018gaussian}.
We argue that the generalized normal assumption also partially reconciles these  two (partially conflicting) research results.

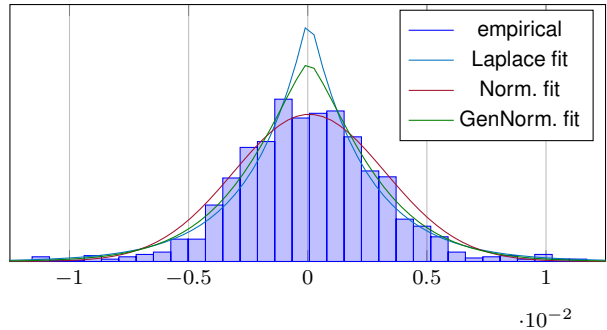
\begin{figure}[t!]
      \centering
     \begin{tikzpicture}[font=\footnotesize\sffamily]
    \definecolor{mycolor1}{rgb}{0.00000,0.44706,0.74118}%
    \definecolor{mycolor2}{rgb}{0.63529,0.07843,0.18431}%
    \definecolor{mycolor3}{rgb}{0.00000,0.49804,0.00000}%
    \begin{axis}[
      height=5 cm,
        width=9.5cm,
         ymin=0,
         ytick=\empty,
         xmin=-0.0125, 
         xmax=0.0125,
         x label style={at={(axis description cs:0.5,-0.1)},anchor=north},
           xmajorgrids={true},
        ]
\coordinate (top) at (axis cs:1,\pgfkeysvalueof{/pgfplots/ymax});

\addplot+[ybar interval,mark=no,draw=blue, fill=blue, fill opacity=0.25]
    table[col sep=comma]{./histogram-data.txt};
    
\addplot+[mark=no,draw=mycolor1 ]
    table[col sep=comma]{./histogram-laplace.txt};
    
\addplot+[mark=no,draw=mycolor2]
    table[col sep=comma]{./histogram-gaussian.txt};
    
\addplot[mark=no,draw=mycolor3]
    table[col sep=comma]{./histogram-gennorm.txt};

 \addlegendimage{green, line legend}
 
\legend{empirical,Laplace fit,Norm. fit,GenNorm. fit}
\end{axis}

\end{tikzpicture}
     \vspace{-0.25cm}
      \caption{Modeling of the gradient updates for layer $42$ -- the bottom layer.}
     \label{fig:grad_modelling_first}
     \vspace{-0.25cm}
  \end{figure}

In Fig. \ref{fig:grad_modelling_first}, we plot (i) the histogram of the DNN gradient updates, (ii) the Laplace  fitting, (iii) the normal fitting and (iv) the GenNorm fitting. As it can observed, the gradient histogram is closer to the GenNorm distribution since it is more concentrated in zero and it has heavier tails than the normal distribution. 

\subsection{DNN Gradient Distortion}
\label{sec:DNN Gradient Distortion}

In choosing the distortion metric for our quantizer design, we keep two considerations in mind: (i) the practitioner perspective on effective  sparsification  and (ii) existing bounds on the accuracy loss from weight perturbation.

In the literature, $\topK$ sparsification  consists in setting all but the $K$ largest weights of the gradient entries to zero at each step of SGD. 
It is well known among the ML practitioners, that $\topK$ sparsification can be effectively used to reduce the dimensionality of the gradient updates while resulting in only a small loss in accuracy 
\cite{alistarh2018convergence,SGD_sparse2,SGD_sparse3}.

Another approach for gradient compression is uniform quantization with non-subtractive dithering \cite{QSGD,SGD_q2,SGD_q4,gandikota2019vqsgd}.
This approach finds its theoretical foundations in works such as \cite[eq. (8)]{lee2020layer} which provides a bound on the loss in accuracy as a function of the $L_2$ perturbation of the DNN weights. 

Given the two compression approaches in which either (i) only the magnitude of the gradients is considered or (ii) the $L_2$ norm of the gradients is used for the quantizer design, we propose a new class of distortion measures combining these approaches:
\ea{
d_{M-L_2}(\gv,\ghv) =  \f 1 d \|g_j\|_M^M \| \gv-\ghv \|_2 
= \frac{1}{d}\sum_{j \in [d]} |g_j|^M  \|g_j-\hat{g}_j\|_2,
\label{eq:our distortion}
}
where $g_j$ and $\hat{g}_j$ denote the $j$-th elements of $\gv$ and $\ghv$ (gradient and quantized vectors), respectively.
In the following, we refer to $d_{M-L_2}$ as the  $M$-magnitude weighted $L_2$ distortion. 
%

\subsection{Quantizer Design}
\label{sec:our metric}
Having substantiated our choice of gradient distribution and the gradient distortion measure, as  in Sec. \ref{sec:DNN Gradient Distribution} and Sec. \ref{sec:DNN Gradient Distortion} respectively, we are now in position to argue for the proposed rate-distortion approach for the quantizer design.
In particular, we argue that the quantizer minimizing the expected $M-L_2$ distortion in  \eqref{eq:our distortion} under the GenNorm assumption provides a good performance in view of  \eqref{eq:G}.
In mathematical language, 
\ea{
\Gsf_{\Rsf}(\whv_T )  \approx \sum_{t \in [T]}  \eta_t \sum_{n \in [n]}d_{M-L_2}(\gv_{tn},\whv_{tn}).
\label{eq:assumption}
}{}
Under the assumption in \eqref{eq:assumption},  the compressor in \eqref{eq:comp} can be designed as the classic rate distortion problem.
%
%
In Sec.  \ref{sec:Numerical Evaluations}, we will consider a scalar quantizer design  minimizing such expected distortion and show its effectiveness in DNN training. 
We note that the $K$-means/LGB algorithm for the class of distortions in \eqref{eq:our distortion} takes a particularly simple form. 
In particular, let $c_k(i)$/$t_k(i)$ be the $i^{\rm th}$ centroid/threshold estimate at iteration $k$ in the scalar LGB algorithms, then
\eas{
c_{k+1}(i+1) & = \frac{\int_{t_{k}(i)}^{t_{k}(i+1)}g^{M+1}\text{pdf}(g)dg}{\int_{t(i)}^{t(i+1)}g^M\text{pdf}(g)dg},\label{eq:Kmeans-updatea}\\
t_{k+1}(i+1) & = \frac{c_{k}(i+1)+c_{k}(i)}{2},\label{eq:Kmeans-updateb}
}{\label{eq:Kmeans-update}}
for $i\in [2^{\Rsf}]$ where $2^{\Rsf}$ is the number of quantization levels, $\text{pdf}(g)$ denotes the distribution fitted to the gradient vector
(in \eqref{eq:Kmeans-update}  we ignore the quantization boundaries due to limited space).  
%


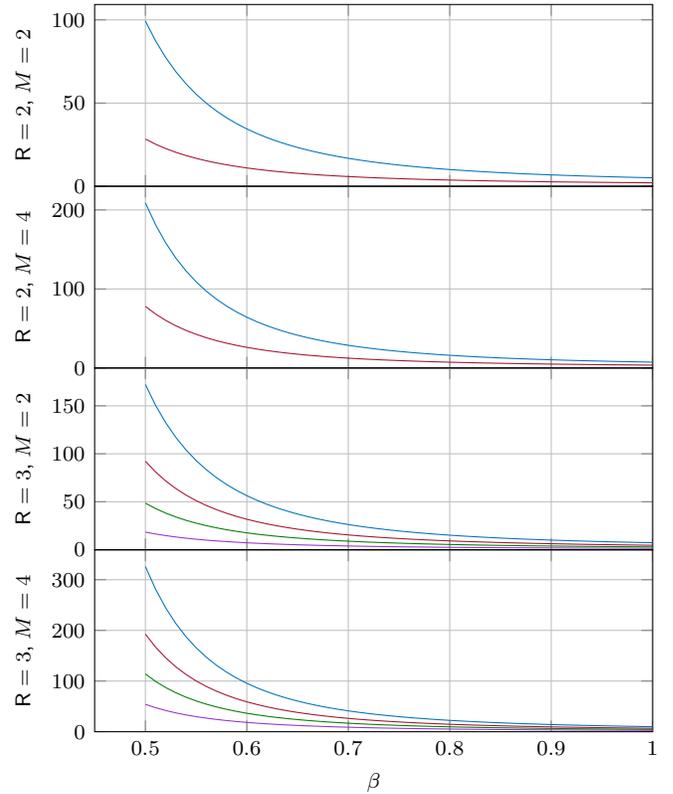
\begin{figure}[t!]
  \centering
 \begin{tikzpicture}[font=\footnotesize\sffamily]
    \definecolor{mycolor1}{rgb}{0.00000,0.44706,0.74118}%
    \definecolor{mycolor2}{rgb}{0.63529,0.07843,0.18431}%
    \definecolor{mycolor3}{rgb}{0.00000,0.49804,0.00000}%
    \definecolor{mycolor4}{rgb}{0.60000,0.19608,0.80000}%
    \definecolor{mycolor5}{rgb}{0.60000,0.19608,0.80000}%
    \definecolor{mycolor6}{rgb}{0.00000,0.49804,0.00000}%
    \definecolor{mycolor7}{rgb}{0.63529,0.07843,0.18431}%
    \definecolor{mycolor8}{rgb}{0.00000,0.44706,0.74118}%
    
    \pgfplotstableread{./K_means_exp_on04_20_2022_17_42_38_R2_num_it1000.txt}{\centRtwo}
    
    \pgfplotstableread{./K_means_exp_on05_16_2022_11_32_08_R2_M4_num_it1000.txt}{\centRtwoMfour}

    \pgfplotstableread{./K_means_exp_on04_21_2022_12_04_28_R3_num_it1000.txt}{\cent}
    
    \pgfplotstableread{./K_means_exp_on05_13_2022_17_03_21_R3_M4_num_it1000.txt}{\centMfour}
        
    \begin{groupplot}[
        group style={
            group name=my plots,
            group size=1 by 4,
            xlabels at=edge bottom,
            xticklabels at=edge bottom,
            vertical sep=0pt,
            horizontal sep=15pt,
              group/every plot/.style={xmin=0,xmax=10},
        },
        height=4 cm,
        width=9 cm,
        ymin=0,
        xmax=1,
        xmajorgrids,
        ymajorgrids,
        legend cell align=left,
     ]


\nextgroupplot[ylabel={$\Rsf =2$, $M=2$}]



\addplot+[draw=mycolor2,no marks]
    table[col sep=comma, x = be, y=c3 ]{\centRtwo};

\addplot+[draw=mycolor1,no marks]
    table[col sep=comma, x = be, y=c4 ]{\centRtwo};


\nextgroupplot[ylabel={$\Rsf =2$, $M=4$}]



\addplot+[draw=mycolor2,no marks]
    table[col sep=comma, x = be, y=c3 ]{\centRtwoMfour};

\addplot+[draw=mycolor1,no marks]
    table[col sep=comma, x = be, y=c4 ]{\centRtwoMfour};


\nextgroupplot[ylabel={$\Rsf =3$, $M=2$}]



 \pgfplotsinvokeforeach{1,...,8}{
  \addplot [draw=mycolor#1]  table[col sep=comma, x = be, y=c#1 ]{\cent};    
  }

\nextgroupplot[ylabel={$\Rsf =3$, $M=4$}, xlabel={$\beta$}]



 \pgfplotsinvokeforeach{1,...,8}{
  \addplot [draw=mycolor#1]  table[col sep=comma, x = be, y=c#1 ]{\centMfour};
}  



\end{groupplot}

\end{tikzpicture}
  \caption{Centroids for the scalar quantizer for the GenNorm distribution as described in Sec. \ref{sec:our metric}.
  }
 \label{fig:centroids}
 \vspace{-0.25cm}
\end{figure}


In Fig. \ref{fig:centroids}, we plot the position of the positive centroids resulting from the $K$-means algorithm in \eqref{eq:Kmeans-update} when applied to the GenNorm distribution with zero mean, unitary variance, and varying $\beta$. 
On the $x$-axis we have the values of $\beta \in [0.5,1]$.
On the $y$-axis we have the position of the positive centroids: since the distribution is symmetric, the centroids are also symmetric. 
On the various panels, we plot four combinations of the quantizer rate -- $\Rsf \in \{2,3\}$ and $M \in \{2,4\}$.
As we can observe, lower values of $\be$ correspond to heavier tail of the distribution which, in turn, result in centroid position further away from zero. 
As $M$ increases, the spreading of the centroid toward large gradient values increases.

\section{Numerical Evaluations}
\label{sec:Numerical Evaluations}
In this section, we compare the performance of our proposed rate-distortion inspired compression with other conventional ML compression techniques, such as floating point conversion and $\topK$ sparsification.  


\subsection{Compression Strategies}
\label{sec:Compression strategies}

Let us begin by describing in further detail the compression techniques we consider in  our simulations.

\noindent
$\bullet$
\textbf{$\topK$ sparsification  + scalar uniform quantization:} For the uniform quantizer with a given quantizer rate $R_{\rm u}$, the $2^{R_{\rm u}}$ quantization centers are uniformly distributed between the minimum and maximum values of the samples in each layer and each iteration.
The sparsification level $K_{\rm u}$ is accordingly chosen such that 
\ea{
d \Rsf = \log \binom{d}{K_{\rm u}} + R_{\rm u} K_{\rm u}. \label{eq:topk1}
}{}

\noindent
$\bullet$
\textbf{$\topK$ sparsification + floating point representation:} In the context of \eqref{eq:comp}, $\topK$ can be applied to meet the rate constraint only once a certain format for representing the gradient entries has been established. So, we consider a \emph{floating point} (fp) representation of the entries with $16$ and $8$ bits. Accordingly, the number of the relationship between the sparsification parameter $K$, the digit precision $p$ (in bits), and the rate constraint $\Rsf$ in \eqref{eq:G}  is
\ea{
d \Rsf = \log \binom{d}{K_{\rm fp}} + K_{\rm fp} p.\label{eq:topk2}
}

\noindent
$\bullet$
\textbf{$\topK$ sparsification +  $2$-magnitude weighted $L_2$:} For our proposed quantizer with a given rate $R_{\rm mw}$, the $2^{R_{\rm mw}}$ quantization centers at each layer and iteration are found by the $K$-means algorithm described in Section~\ref{sec:our metric} for $M=2$. We use a $\topK$ sparsification before our compressor where the sparsification level $K_{\rm mw}$ satisfies the following:
\ea{
d \Rsf = \log \binom{d}{K_{\rm mw}} + K_{\rm mw} R_{\rm mw}.\label{eq:topk3}
}

\subsection{Simulation Results}
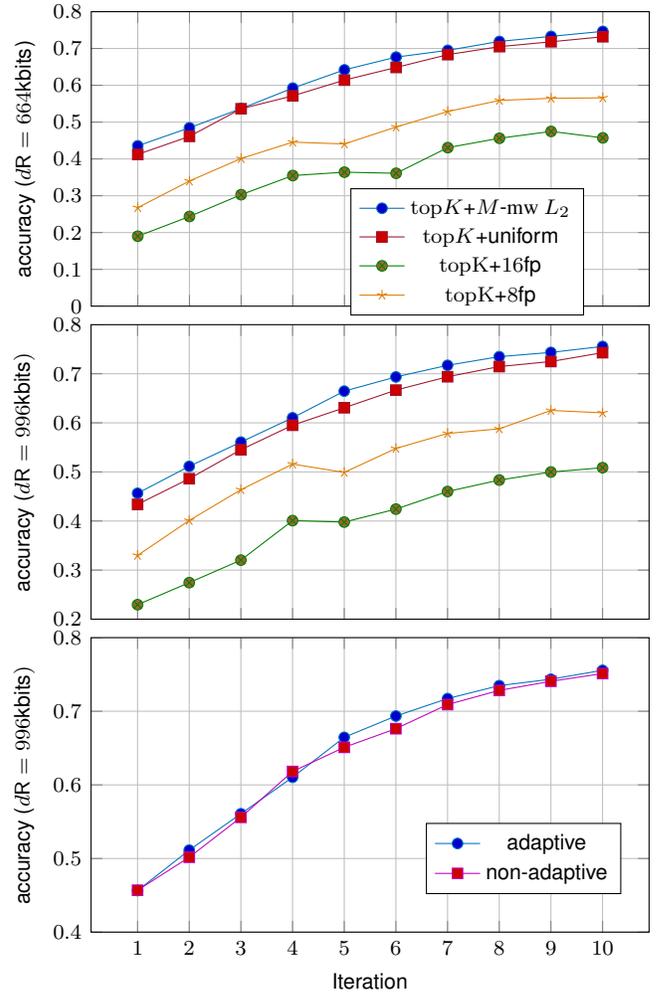
\begin{figure}[t!]
  \centering
 \begin{tikzpicture}[font=\footnotesize\sffamily]
\definecolor{mycolor1}{rgb}{0.00000,0.44706,0.74118}%
\definecolor{mycolor2}{rgb}{0.63529,0.07843,0.18431}%
\definecolor{mycolor3}{rgb}{0.00000,0.49804,0.00000}%
\definecolor{mycolor4}{rgb}{0.87059,0.49020,0.00000}%
\definecolor{mycolor5}{rgb}{0.00000,0.44700,0.74100}%
\definecolor{mycolor6}{rgb}{0.74902,0.00000,0.74902}%

    \begin{groupplot}[
        group style={
            group name=my plots,
            group size=1 by 3,
            xlabels at=edge bottom,
            xticklabels at=edge bottom,
            vertical sep=7 pt,
            horizontal sep=5pt
        },
        height=5.5cm,
        width=9cm,
        xmajorgrids,
        ymajorgrids,
     ]
]


\nextgroupplot[ylabel={accuracy ($d\mathsf{R}=664\text{kbits}$)},ymin=0,ymax=0.8,
ytick distance=0.1,xtick={1,2,3,4,5,6,7,8,9,10},legend to name={CommonLegend},legend style={legend columns=1}]
\coordinate (top) at (axis cs:1,\pgfkeysvalueof{/pgfplots/ymax});


\addplot+[draw=mycolor1]
    table[col sep=comma]{./accuracy-k-means-R2-fixed-seed0.txt};
\addplot+[draw=mycolor2]
    table[col sep=comma]{./accuracy-uniform-scalar-R2-fixed-seed0.txt}; 
\addplot+[draw=mycolor3]
    table[col sep=comma]{./accuracy-topK-16fp-R2-seed0.txt}; 
\addplot+[draw=mycolor4]
    table[col sep=comma]{./accuracy-topK-8fp-R2-seed0.txt};

\legend{$\topK$+$M$-{\rm mw} $L_2$,$\topK$+uniform,$\rm \topK$+$16$fp,$\rm \topK$+$8$fp}


\nextgroupplot[ylabel={accuracy ($d\mathsf{R}=996\text{kbits})$},ymin=0.2,ymax=0.8,
ytick distance=0.1,xtick={1,2,3,4,5,6,7,8,9,10}
]
\addplot+[draw=mycolor1]
    table[col sep=comma]{./accuracy-k-means-R3-fixed-seed0-M4.txt};
\addplot+[draw=mycolor2]
    table[col sep=comma]{./accuracy-uniform-scalar-R3-fixed-seed0.txt}; 
\addplot+[draw=mycolor3]
    table[col sep=comma]{./accuracy-topK-16fp-R3-seed0.txt}; 
\addplot+[draw=mycolor4]
    table[col sep=comma]{./accuracy-topK-8fp-R3-seed0.txt};

\nextgroupplot[ylabel={accuracy ($d\mathsf{R}=996\text{kbits})$},xlabel={Iteration},ymin=0.4,ymax=0.8,
ytick distance=0.1,xtick={1,2,3,4,5,6,7,8,9,10},
legend style={at={(0.6,0.25)},anchor=west}
]
\addplot+[draw=mycolor1]
    table[col sep=comma]{./accuracy-k-means-R3-fixed-seed0-M4.txt};

\addplot+[draw=mycolor6,]
    table[col sep=comma]{./fixed-beta-accuracy-k-means-R3-seed0.txt};

\legend{adaptive, non-adaptive}

\end{groupplot}

\path (5,-0.5) -- node[above]{\ref{CommonLegend}} (5,0);
\end{tikzpicture}
  \caption{Testing accuracy as a function of the iteration number for $d\Rsf=664\text{kbits}$  and $d\Rsf = 996\text{kbits}$, respectively. In the former case we choose $M=2$, in the latter $M=4$.}
 \label{fig:accuracy}
 \vspace{-0.25cm}
\end{figure}

In Fig.~\ref{fig:accuracy}, we plot the accuracy of the trained network for different quantization rates $d\Rsf=664\text{kbits}$ and $d\Rsf=996\text{kbits}$. We use the network introduced in Table~\ref{tab:model_layers}.
 The compression strategies are as detailed in  Sec. \ref{sec:Compression strategies}. For $d\Rsf=664\text{kbits}$, we choose the following parameters for each of the quantizers:
 \begin{itemize}
     \item $\topK$  + uniform: $R_{\rm u}=2$, $K_{\rm u}=331725$,
     \item $\topK$ + 16fp: $p=16$, $K_{\rm fp}=41466$,
     \item $\topK$ + 8fp : $p=8$, $K_{\rm fp}=82932$,
     \item $\topK$ + $2$-{\rm mw} $L_2$: $M=2$, $R_{\rm mw}=2$, $K_{\rm mw}=331725$,
 \end{itemize}
 and for $d\Rsf=996\text{kbits}$, the following set of parameters are proposed:
  \begin{itemize}
     \item $\topK$  + uniform: $R_{\rm u}=3$, $K_{\rm u}=331725$,
     \item $\topK$ + 16fp: $p=16$, $K_{\rm fp}=62199$,
     \item $\topK$ + 8fp : $p=8$, $K_{\rm fp}=124398$,
     \item $\topK$ + $4$-{\rm mw} $L_2$: $M=4$, $R_{\rm mw}=3$, $K_{\rm mw}=331725$.
 \end{itemize}
 The blue dotted curve shows the accuracy of our proposed quantizer versus the iteration number.  The performances of the uniform compressor, $\topK$-8fp and $\topK$-16fp are shown in red squared, yellow starred, green circled curves, respectively. 

 We  point out that the optimization over $M$ in the first two panels was only partial: further tuning of this parameter would result in further performance improvements.
 Also note that the quantizer for {\rm mw} $L_2$ and uniform quantization  is chosen adaptively. In other words, the quantizer is chose as a function of the empirical gradient distribution at each iteration. 
 For the case of {\rm mw} $L_2$, this is attained by pre-calculating the centroid for different values of $\be$ as in Fig. \ref{fig:centroids}.
 At each iteration, then, the gradient vector is pre-processed to obtain a vector  of zero mean and unitary variance which is then quantized using the pre-calculated scalar quantizer. 
  
\smallskip

\noindent
 {\bf Adaptive vs. non-adaptive {\rm mw} $L_2$  quantization:}
 In the first two panels of Fig.~\ref{fig:accuracy} we consider an adaptive quantization strategies. That is, the quantizer is chosen at each iteration as a function of the empirical gradient distribution. 
 In actuality, the GenNorm modelling of \cite{chen2021dnn} produces rather stable estimates of the GenNorm parameters describing the empirical gradient distribution. 
 For this reason, one can consider the performance when the quantizer is non-adaptive. 
 The performance of such a non-adaptive quantizer, together with its adaptive counterpart, is depicted in the third panel of Fig. \ref{fig:accuracy}.
 The small loss of performance from a non-adaptive quantizer suggests that, in a practical scenario, one would do a first boot-up training phase in which the gradient parameters are estimated. The quantizer would then be chosen as a function of such estimates and used throughout the remaining iterations. 
 

%

\smallskip

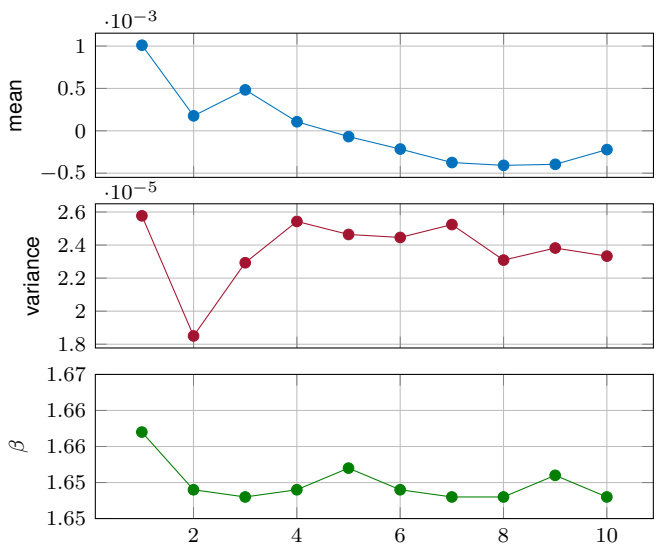
\begin{figure}[t!]
      \centering
      \vspace{+0.25cm}
     \begin{tikzpicture}[font=\footnotesize\sffamily]
    \definecolor{mycolor1}{rgb}{0.00000,0.44706,0.74118}%
    \definecolor{mycolor2}{rgb}{0.63529,0.07843,0.18431}%
    \definecolor{mycolor3}{rgb}{0.00000,0.49804,0.00000}%
    \definecolor{mycolor4}{rgb}{0.60000,0.19608,0.80000}%
    \definecolor{mycolor5}{rgb}{0.60000,0.19608,0.80000}%
    \definecolor{mycolor6}{rgb}{0.00000,0.49804,0.00000}%
    \definecolor{mycolor7}{rgb}{0.63529,0.07843,0.18431}%
    \definecolor{mycolor8}{rgb}{0.00000,0.44706,0.74118}%
    
    \pgfplotstableread{./K_means_exp_on04_20_2022_17_42_38_R2_num_it1000.txt}{\centRtwo}
    
    \pgfplotstableread{./K_means_exp_on05_16_2022_11_32_08_R2_M4_num_it1000.txt}{\centRtwoMfour}

    \pgfplotstableread{./K_means_exp_on04_21_2022_12_04_28_R3_num_it1000.txt}{\cent}
    
    \pgfplotstableread{./K_means_exp_on05_13_2022_17_03_21_R3_M4_num_it1000.txt}{\centMfour}
        
    \begin{groupplot}[
        group style={
            group name=my plots,
            group size=1 by 3,
            xlabels at=edge bottom,
            xticklabels at=edge bottom,
            vertical sep=10pt,
            horizontal sep=15pt,
              group/every plot/.style={xmin=0,xmax=10},
        },
        height=3.5 cm,
        width=9 cm,
        xmajorgrids,
        ymajorgrids,
        legend cell align=left,
     ]


\nextgroupplot[ylabel={mean}]

\addplot+[draw=mycolor1,mark options={fill=mycolor1}]
    table[col sep=comma]{./mean-kmeans-R3-seed0.txt};


\nextgroupplot[ylabel={variance}]

\addplot+[draw=mycolor2,,mark options={fill=mycolor2}]
    table[col sep=comma]{./variance-k-means-R3-seed0.txt};


\nextgroupplot[ylabel={$\be$},ymin=1.65,ymax=1.67]

 \addplot+[draw=mycolor3,mark options={fill=mycolor3}]
     table[col sep=comma]{./beta-k-means-R3-seed0.txt};

\end{groupplot}

\end{tikzpicture}
      \caption{Modeling of the gradient updates for layer $24$ -- the middle layer for the first ten iterations.}
     \label{fig:grad_modelling}
  \end{figure}

\noindent
 {\bf GenNorm Modelling under {\rm mw} $L_2$  quantization:}
  As a last numerical consideration, we would like to argue that the GenNorm modelling of the DNN gradients remain valid also under {\rm mw} $L_2$  quantization. 
  In other words, the choice of quantization strategy does not fundamentally alters the gradient distribution. 
  Indeed, in Fig.~\ref{fig:grad_modelling}, we plot the GenNorm parameters modelling the last layer of the DNN and observe the consistency of these estimate through the iteration.

\smallskip

The code for the numerical evaluations in this section is provided online at \url{https://github.com/sadafsk/FL\_RD}.

\section{Conclusion}
In this paper, the problem of efficient gradient compression for DNN has been considered. 
In particular, we approach this problem from a rate-distortion perspective in which the quantizer used for compression is designed under an assumption on (i) the distribution of the gradient updates and (ii) the distortion measure which minimized the loss in accuracy. 
For (i), we assume that the gradient updates are in each DNN layer and at each iteration are identically and independently distributed according to a generalized normal distribution.
For (ii), we assume that the distortion capturing the relationship between gradient perturbation and loss in accuracy is the $M$-magnitude weighted $L_2$ norm, that is the $L_2$ norm in which the error is further multiplied by the magnitude of the gradient to the power of $M$. 
We argue that this choice of distortion naturally bridges between two classical gradient sparsification approaches: $M=0$ recovers uniform quantization, while $M \goes \infty$ recovers $\topK$ sparsification.  
In this work, both of these assumptions are validated through numerical evaluations: a more theoretical justification of these two assumptions will be investigated in our future research. 
Furthermore, in our simulation the optimization over the weighting parameter $M$ was only partial: a more detailed investigation over the role of this parameter is also left for future research.










\bibliographystyle{IEEEtran}
\bibliography{FL_lattice}

\end{document}